\documentclass[10pt,twocolumn,letterpaper]{article}

\usepackage{wacv}
\usepackage{times}
\usepackage{epsfig}
\usepackage{graphicx}
\usepackage{amsmath}
\usepackage{amssymb}
\usepackage{url}
\usepackage{booktabs}
\usepackage[ruled,linesnumbered]{algorithm2e}
\SetKwComment{Comment}{//}{}

\def\wacvPaperID{12} 

\wacvfinalcopy 

\ifwacvfinal
\fi

\ifwacvfinal
\usepackage[breaklinks=true,bookmarks=false]{hyperref}
\else
\usepackage[pagebackref=true,breaklinks=true,colorlinks,bookmarks=false]{hyperref}
\fi

\begin{document}

\title{PP-HumanSeg: Connectivity-Aware Portrait Segmentation \\ with a Large-Scale Teleconferencing Video Dataset}

\author{Lutao Chu, Yi Liu, Zewu Wu, Shiyu Tang, Guowei Chen, Yuying Hao \\ Juncai Peng, Zhiliang Yu, Zeyu Chen, Baohua Lai, Haoyi Xiong \\

{Baidu, Inc.}

}

\maketitle

\begin{abstract}
   As the COVID-19 pandemic rampages across the world, the demands of video conferencing surge. To this end, real-time portrait segmentation becomes a popular feature to replace backgrounds of conferencing participants. While feature-rich datasets, models and algorithms have been offered for segmentation that extract body postures from life scenes, portrait segmentation has yet not been well covered in a video conferencing context. To facilitate the progress in this field, we introduce an open-source solution named PP-HumanSeg. This work is the first to construct a large-scale video portrait dataset that contains 291 videos from 23 conference scenes with 14K fine-labeled frames and extensions to multi-camera teleconferencing. Furthermore, we propose a novel \emph{Semantic Connectivity-aware Learning (SCL)} for semantic segmentation, which introduces a semantic connectivity-aware loss to improve the quality of segmentation results from the perspective of connectivity. And we propose an ultra-lightweight model with SCL for practical portrait segmentation, which achieves the best trade-off between IoU and the speed of inference. Extensive evaluations on our dataset demonstrate the superiority of SCL and our model. The source code is available at \href{https://github.com/PaddlePaddle/PaddleSeg}{https://github.com/PaddlePaddle/PaddleSeg}.
\end{abstract}

\section{Introduction}

\begin{figure}
\centerline{\includegraphics[width=0.48\textwidth]{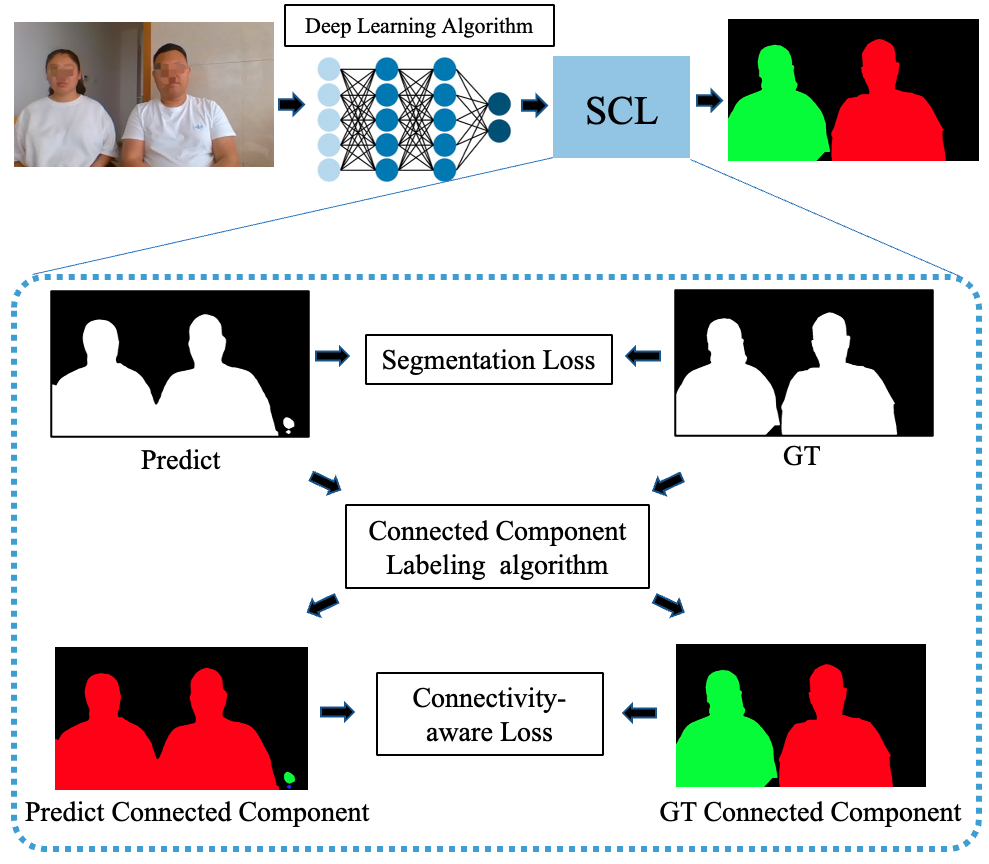}}
\caption{An illustration of the proposed Semantic Connectivity-aware Learning (SCL) approach for semantic segmentation, which improves segmentation performance from the perspective of connectivity.}
\label{framework}
\end{figure}

Portrait segmentation~\cite{shen2016automatic} has brought great success in various entertainment applications, such as virtual background, beautifying filters, character special effects. Among these applications, video conferencing has become a major scenario for portrait segmentation, where participants could automatically replace their private backgrounds (e.g., ones from private rooms) with virtual scenes. 

The outbreak of coronavirus has further accelerated the prevalence of video conferencing to dramatically replace the traditional face-to-face meetings, as working-from-home has been desired~\cite{sander2020coronavirus}. Moreover, compared to the traditional video conferencing that links participants from different offices/conference rooms, the current meeting scenes become much more diverse in surroundings and lighting conditions, as live videos are recorded from each participant's home. Participants may show various postures and actions, and even wear face masks. In addition, participants sometimes access to the teleconferencing using a thin client, such as a webpage for chat based on JavaScript running on a browser, or a chat App running on mobile devices. Thus, there frequently needs to serve portrait segmentation tasks in resource-limited computing platforms (e.g., webpages and smartphones without powerful GPUs) while ensuring real-time performance for teleconferencing on-the-air. All these practical issues in post-COVID-19 video conferencing have brought great challenges and opportunities to the portrait segmentation field.

\begin{figure*}[t!]
\centerline{\includegraphics[width=0.95\textwidth]{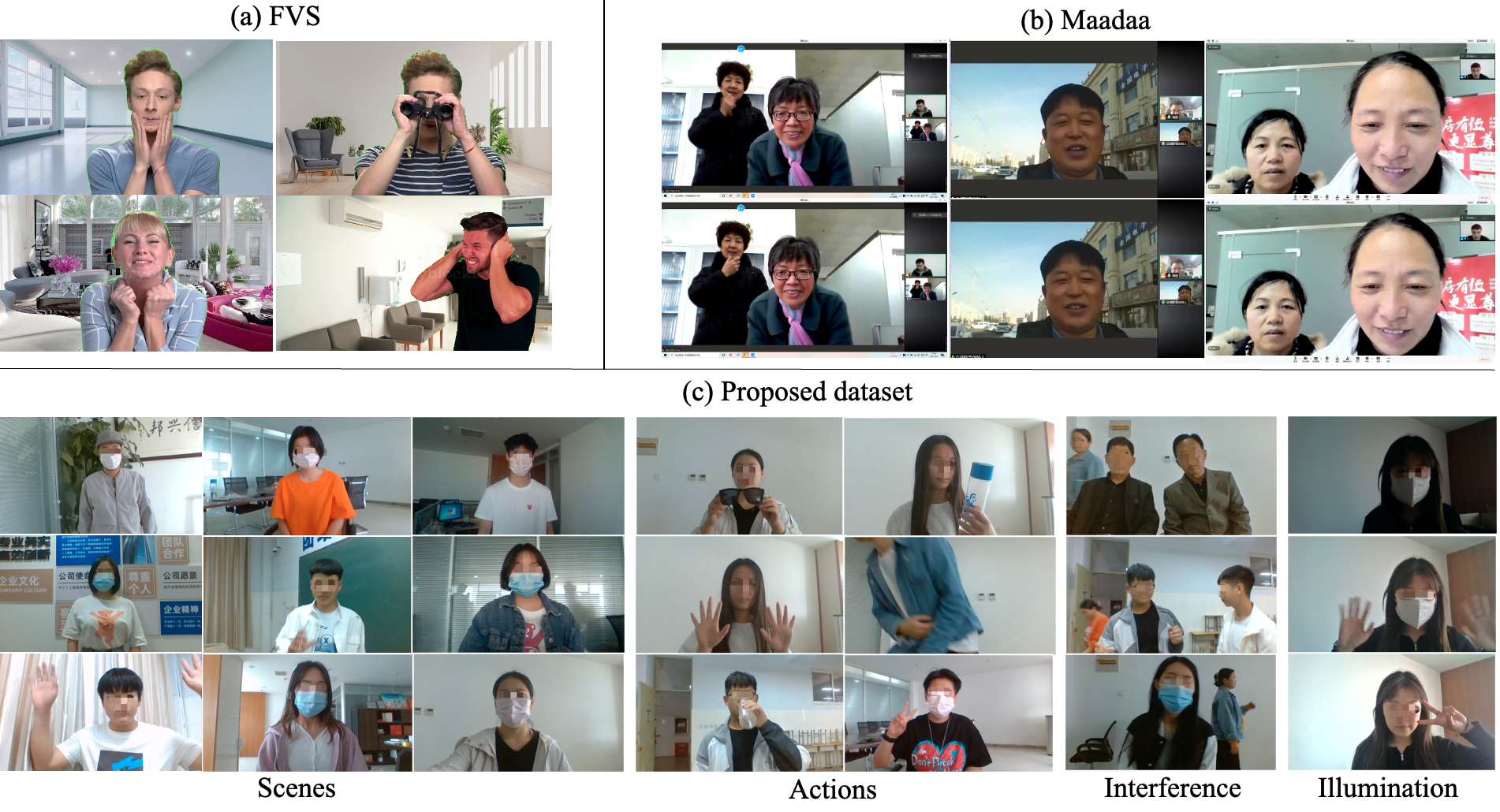}}
\caption{Examples of our dataset and existing datasets. (a) FVS contains only 4  green-screen videos. Due to the composition effect, the labels are not smooth enough. (b) Maadaa contains a lot of similar images and irrelevant information of the interface of the software, e.g. virtual buttons, small windows. (c) The proposed dataset contains various teleconferencing scenes, various actions of the participants, interference of passers-by and illumination change. Note that all videos with human subjects in the proposed datasets have granted the rights to use and disseminate for scientific research purposes.}
\label{dataset}
\end{figure*}

Actually, many works have been done on both datasets and methodologies for portrait segmentation. For datasets, there are EG1800~\cite{shen2016automatic}, AISeg~\cite{aiseg}, FVS~\cite{DBLP:journals/corr/abs-2104-09752}, Maadaa~\cite{maadaa}, as shown in figure \ref{dataset}.
However, they are rarely applied for video conferencing tasks. The datasets for video conferencing either are with low picture quality and high redundancy or even contain synthesized images. \emph{Thus, a new dataset with real-world teleconferencing videos of high picture quality and fine-grained labels is required}.


In terms of segmentation methods, a great number of works have been proposed to address context information~\cite{yuan2020object,zhao2017pyramid,chen2018encoder}, multi-scale adaptation~\cite{chen2016attention,tao2020hierarchical}, fine edge processing~\cite{kirillov2020pointrend,yuan2020segfix,cheng2020boundary,cheng2021boundary,hao2021edgeflow}, category imbalance loss~\cite{berman2018lovasz,milletari2016v} issues. However, these approaches are designed for generic semantic segmentation yet not optimized for portrait segmentation. Although portrait segmentation is a sub-type of semantic segmentation, it has distinct characteristics comparing with other object segmentation. The person can be regarded as a non-rigid object, so that the postures and appearances are varying, which is challenging in the semantic segmentation task. In addition, generic semantic segmentation is pixel-level classification which ignores the completeness of person instances. \emph{Thus, a new learning approach that takes care of completeness of person instances, subject to varying human actions/postures, is required for portrait segmentation in teleconferencing.}


In addition, to achieve portrait segmentation on mobile devices, several lightweight models for semantic segmentation have been proposed~\cite{BMVC2019,wang2020deep}. However, the results of these models~\cite{BMVC2019,wang2020deep} evaluated on portrait dataset are unsatisfactory. \emph{Thus, a lightweight model that could deliver real-time portrait segmentation on resource-limited platforms (e.g., mobile devices and browsers) is required.}



Therefore, we introduce an open-source solution for practical portrait segmentation named PP-HumanSeg. In this work, we construct a large-scale video portrait dataset including 291 meeting videos in 23 different scenes. To facilitate researchers in the field, we provide 14,117 fine-annotated images. To improve the completeness of person instances, we propose a new Semantic Connectivity-aware Learning (SCL) approach, where the connected component concept is used to represent the completeness of the person. The proposed approach improves the consistent connectivity between the segmentation results and the ground truth. Finally, we propose an ultra lightweight segmentation network using SCL, which achieves the best trade-off among mIoU and the inference speed. The contributions of this paper are as follows:
\begin{itemize}
\item We release a large-scale video portrait dataset that contains 291 videos from 23 conference scenes with 14K fine-labeled frames provided, to facilitate the progress in portrait segmentation in video conferencing. Please refer to figure~\ref{dataset} for comparisons with existing datasets, such as FVS~\cite{DBLP:journals/corr/abs-2104-09752} and Maadaa~\cite{maadaa}.

\item We propose a novel Semantic Connectivity-aware Learning (SCL) framework for portrait segmentation, which improves segmentation performance from the perspective of connectivity. 

\item We propose an ultra-lightweight model with SCL for practical portrait segmentation, which achieves the best trade-off between performance and the inference speed. Extensive evaluations on our dataset demonstrate the superiority of SCL and our model. 
\end{itemize}
To the best of our knowledge, it is the first video portrait dataset with various scenes, character appearances and actions for video conferencing, with non-trivial baseline models/algorithms offered.

\section{Related Works}
While the main contributions of this paper include a new dataset, a new learning framework, and a new lightweight model all for portrait segmentation in teleconferencing setting, we thus introduce and discuss the related works from these three perspectives. 

\paragraph{Datasets.} There are several popular portrait datasets, such as EG1800~\cite{shen2016automatic},  FVS~\cite{DBLP:journals/corr/abs-2104-09752}, Maadaa~\cite{maadaa} and AISeg~\cite{aiseg}. Compared to EG1800~\cite{shen2016automatic}, AISeg~\cite{aiseg}, and FVS~\cite{DBLP:journals/corr/abs-2104-09752} that provided (self-)portrait images and segmentation labels of persons under various indoor/outdoor or even virtual backgrounds, our work offers massive fine-labeled frames of real-world videos for teleconferencing. Maadaa~\cite{maadaa} also provided images collected from video conferencing scenarios, but they were all screenshots from the video conferencing applications that incorporate irrelevant and noisy pixels, such as software interfaces. In addition, all existing datasets do not include persons wearing face masks, which is unavoidable for post-COVID-19 teleconferencing. 

\paragraph{Learning Methods and Lightweight Models.} The existing learning algorithms for semantic segmentation mainly incorporate cross entropy loss, lovasz loss~\cite{berman2018lovasz}, dice loss~\cite{milletari2016v}, and RMI loss~\cite{zhao2019region} for training. In addition, upon these training methods, the multi-branch networks have been proposed to improve the lightweight models~\cite{yu2018bisenet,BMVC2019,poudel2018contextnet,mazzini2018guided} for generic segmentation problem.  Compared to these works, we propose a new SCL framework that incorporates a new loss, namely \emph{semantic connectivity-aware loss}, to improve the completeness of segmentation results for person instances and introduce a new model design, namely \emph{ConnectNet} to facilitate ultra-lightweight connectivity-aware portrait segmentation. 

Note that some face-related libraries, such as~\cite{wang2021face,zhao2018towards}, also include face detection modules that can improve the performance of portrait segmentation. Due to the page limits, we do not include the discussion on them here.

\section{The Proposed Dataset}
In this section, we introduce the ways we collect and label images and videos for portrait segmentation in real-world teleconferencing settings.
\subsection{Data Collection} 

In order to get closer to the real video conference data distribution, we collect the videos in 23 common conference scenes including meeting rooms, offices, public office areas, living room, classrooms, etc. In addition, the participants perform various actions, e.g. waving hands, getting up and sitting down, drinking water, using mobile phones, shaking, etc. We also collected a large number of pictures of people wearing masks. Finally, we get a large-scale dataset of 291 videos with 1280x720 resolution. In order to reduce redundancy, we extract frames from the videos at a low frame rate of 2.5 FPS to get 14117 HD images. The diversity of collected images is shown in figure~\ref{dataset}(c). 

\subsection{Data Labeling}

We recruited several professional annotators to label the collected data. They provide high-quality labels of our dataset in both pixel level and video level. 

\subsubsection{Pixel-Level Labeling}

In fact, the annotation of portrait segmentation usually has two ambiguous instances, 1) hand-held items, 2) distant passerby or people with backs. The annotation of them depends on the practical applications, as well as the definition of foreground and background. In video conferencing, the purpose of portrait segmentation is highlighting participant-related parts rather than the surroundings. The hand-held items highly related to the activities of participants, such as mobile phone, glasses and cup. However, distant passerby or people with backs are not participants of the video conference, which should be ignored. Therefore, all hand-held items are labelled together with human body. Distant passers-by or people with backs are not labelled, even though they are usually labelled in other applications of portrait segmentation. 
	
\subsubsection{Video-Level Labeling}

Following the practice of VOC~\cite{everingham2010pascal} and PSS~\cite{zhang2021personalized}, we annotate our videos based on the objects appeared in the video. Each video clip has multi-class attributes, e.g. the scene id, the number of participants, the activity of participants, wearing face mask, passers-by. The video-level annotation can be used to video description and multi-task learning, which also provides a good starting point to human activity analysis study in video conferencing.

\subsection{Video Synthesis for Teleconferencing}

\begin{figure}[t!]
\centerline{\includegraphics[width=0.48\textwidth]{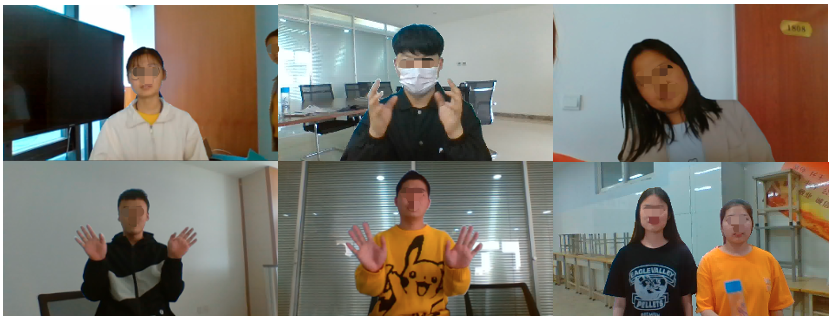}}
\caption{Examples of composited videos for teleconferencing.}
\label{composition}
\end{figure}

Besides the 14K fine-labeled images, we also collected pure-background images in 90 different video conferencing scenes. Then we use a simple video composition strategy to augment the dataset further. The high-quality portrait masks allow us to extract the portrait parts precisely, and much more labeled images is composed of the extracted portrait parts and pure-background images. Through data composition, we generate around one million images eventually. Due to high-quality annotation, the edges of the composition data are smooth and look natural, as shown in figure~\ref{composition}.

\section{SCL: Semantic Connectivity-aware Learning for Portrait Segmentation}

In this section, we present the design of Semantic Connectivity-aware Learning (SCL) framework (shown in figure \ref{framework}) for semantic segmentation. To improve the completeness of segmentation results, we define a new concept namely \emph{semantic connectivity} to represent the portrait segmentation results and ground truth. Specifically, in addition to using traditional semantic labels as supervision, SCL extracts the connected components from semantic labels and uses them as the supervision signal via a Semantic Connectivity (SC) loss. Note that SCL framework complement with other deep neural architectures (e.g. CNNs, Transformer-based Networks~\cite{zheng2021rethinking,liu2021swin,xie2021segformer,vit}) to boost the performance of portrait segmentation.

\subsection{Semantic Connectivity between Components in Segmentation}
\begin{figure}
\centerline{\includegraphics[width=0.48\textwidth]{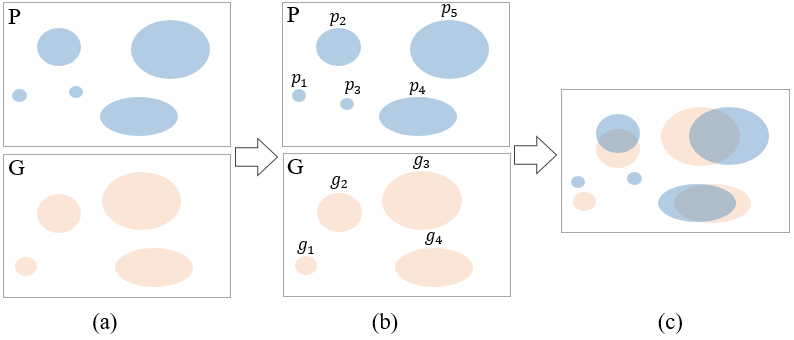}}
\caption{Connected components calculation and matching. (a) It indicates prediction and ground truth, i.e. $P$ and $G$. (b) Connected components are generated through the CCL algorithm~\cite{grana2010optimized}, respectively. (c) Connected components are matched using the IoU value.}
\label{concept}
\end{figure}

In this work, we use the connected components to represent the completeness of the portrait segmentation. In topology, connected component is a maximal subset of a topological space that cannot be covered by the union of disjoint subsets. In portrait segmentation, we take the region of a person instance as a connected component. Figure~\ref{concept} shows an example for connected components calculation and matching.

We find the connected components of predictions ($P$) and ground truth ($G$), respectively. Connected components calculation is a fundamental principle in image processing, where there are many methods, e.g. connected component labelling (CCL) and edge thinning. In our approach, we use a CCL algorithm to calculate the connected components, because of its robustness~\cite{grana2010optimized}. We then traverse all connected components of $G$ and $P$ to find all pairs that intersect with each other. In figure~\ref{concept}, there are three pairs, i.e. $[g_2, p_2]$, $[g_3, p_5]$, $[g_4, p_4]$, and three isolated components, i.e. $p_1, p_3, g_1$. Note that a connected component in $G$ could have intersections with multiple connected components in $P$, which is not be indicated in the figure.

Assuming $g_i$ is paired with $\{p_1, p_2, ..., p_k\}$, the connectivity of $g_i$ is denoted as $C_i$, which is calculated with the equation as follows. 
\begin{equation}
\label{eq:connectivity}
C_i(P) = \frac{1}{k} \Sigma_{k=1}^{k}\mathrm{IoU}(g_i, p_k)\in (0, 1]
\end{equation}
\begin{equation}
\label{eq:iou}
\mathrm{IoU}(g_i, p_k) = \frac{|g_i\cap p_k|}{|g_i\cup p_k|}\ 
\end{equation}
In particular, when $g_i$ is only paired with one connected component in $P$, e.g. $p_j$, $C_i$ equals to IoU between $g_i$ and $p_j$. If $g_i$ is an isolated component, $C_i$ equals to 0.

Finally, we define the semantic connectivity (SC) of the entire image given the graph of components in the ground truth $G$ and the graph in the prediction $P$ as the follow.
\begin{equation}
\label{eq:sc}
 \mathrm{SC}(P,G) = \frac{1}{N} \Sigma_{i=1}^N C_i(P)
\end{equation}
where $N$ is the total number of both pairs and isolated components. Note that for $\forall P,G$ we have $\mathrm{SC}(P,G)\in [0, 1]$. 

\subsection{Learning with SC Loss}

To enable the semantic connectivity-aware learning, the SCL frameworks uses a novel loss function based on the proposed semantic connectivity, which minimize the inconsistency of connectivity between the prediction and the ground truth. In addition, when no intersection between the prediction and the ground truth, we use an area-based loss function to better optimize the model. 


The mathematical notation is the same as in the previous section, we denote Semantic Connectivity-aware (SC) Loss as $L_\mathrm{SC}$. If there is at least a pair between $P$ and $G$, $L_\mathrm{SC}$ is defined as follow.
\begin{equation}
 L_\mathrm{SC}(P,G) = 1-\mathrm{SC}(P,G)\ ,
\end{equation}
where for $\forall P,G$ we have $L_\mathrm{SC}(P,G)\in [0, 1]$. 

Note that there is a special case that no pair exists between $P$ and $G$, and connectivity is becoming to be 0. It could happen in the beginning of training, due to random initialization of parameters. However, 0-connectivity in SCL would lead to zeros gradients, and the weights cannot not be updated accordingly the connectivity. For such special case, we design a non-trivial loss function to cold start the process. Specifically, to ensure the continuity and differentiability of the loss function in the cold-start phase, we write the SC loss $ L_\mathrm{SC}$ as follow.
%
%
\begin{equation}
L_{SC}(P, G) = \frac{|P\cup G|}{|I|},\label{lsc2}
\end{equation}
where $I$ represents the image and $|\cdot|$ represents the area of the region (total number of pixels in the region), and for $\forall P, Q$ we have $L_{SC}(P, G) \in (0, 1]$. 


Finally, SCL incorporates the SC Loss as a regularizer to complement with the segmentation losses (denoted as $L_S$. e.g. cross entropy loss) in the form of $L=L_S + \lambda * L_\mathrm{SC}$ to optimize the model. The hyper-parameter $\lambda$ denotes a weight to make trade-off between the SC loss and the segmentation loss.

\section{ConnectNet: an Ultra-lightweight Neural Network for Portrait Segmentation}

\begin{figure}
\centerline{\includegraphics[width=0.48\textwidth]{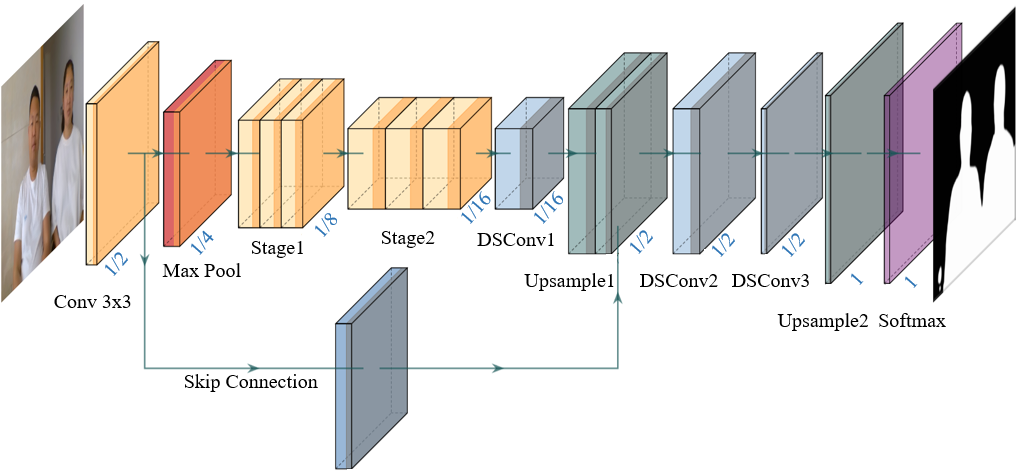}}
\caption{ConnectNet: an Ultra-lightweight Model for Portrait Segmentation.}
\label{connectnet}
\end{figure}

We propose an ultra-lightweight segmentation network to work with SCL, namely \emph{ConnectNet}, as shown in figure~\ref{connectnet}. ConnectNet adopts an encoder-decoder structure. The encoder follows an inverted bottleneck block~\cite{ma2018shufflenet} design with channel-shuffle operation to extract features efficiently.  To reduce the computation loads while  maintaining high resolutions, ConnectNet compresses the number of stages and channels, where every stage is stacked by multiple inverted bottleneck blocks. Moreover, ConnectNet incorporates depth-wise separable convolution to improve the decoding efficiency in the decoder, where depth-wise separable convolution decomposes the ordinary convolution into depth-wise convolution and point-wise convolution so as to further reduce computation loads. 

With features extracted in an encoder-decoder network with bottleneck layers, the encoder would lower the resolution of the feature map and lose the spatial details. Spatial information is critical in segmentation tasks. Therefore, the proposed network connects the encoder and decoder across layers through a skip connection to integrate the underlying texture features, which is more conducive to generating fine masks. At the same time, the skip connection directly reuses the features extracted by the encoder without additional computation costs. 

\section{Experiments}

\subsection{Experiment settings}
All of our experiments are conducted on two Tesla V100 GPUs of 32GB using PaddlePaddle\footnote{https://github.com/PaddlePaddle/Paddle}~\cite{paddlepaddle}. Code and pretrained models are available at PaddleSeg\footnote{https://github.com/PaddlePaddle/PaddleSeg}~\cite{liu2021paddleseg}. During training, we use polynomial decay with power equal to 0.9, and the learning rate equals to 0.05 and 0.025 for HRNet-W18-small and other networks respectively. We use SGD as our optimizer with weight decay parameter being 0.0005. We apply data augmentation methods including scale, crop, flip, and color distortion for training.  We use BBDT algorithm~\cite{grana2010optimized} for connected component labeling. 

In order to avoid similar images in the validation set and test set, we divide the dataset by scene level. The proposed dataset is randomly divided into a training set with 11 scenes and 9006 images, a validation set with 6 scenes and 2549 images, and a test set with 6 scenes and 2562 images.  We train our model with the batch size of 128. For all experiments, we take mIoU and pixel accuracy as evaluation metrics. 

\subsection{Experiment Results}

\subsubsection{Hyper-parameter}
SCL optimizes the network using a weighted combination of cross entropy loss and SC loss. Different combination coefficient may bring different effects. In order to show that the SC loss is parameter in-sensitive and robust. We conduct 5 experiments with different weight coefficients, i.e. 0.01, 0.05, 0.1, 0.5, and 1.0. 

As shown in Table~\ref{different_ratios}, the mIoUs under almost all coefficients are improved on different combination of SC loss and segmentation loss. We set $\lambda$ as 1.0 in the following experiment settings. 


\begin{table}[ht]
\centering
\resizebox{0.47\textwidth}{!}{
\begin{tabular}{l|cccccc}
 \toprule 
Weight coefficient & baseline & 0.01 & 0.05 & 0.1 & 0.5 & 1.0\\ \toprule
mIoU & 93.0 & 94.2 & 93.9 & 94.5 & 92.6 & \textbf{94.6}\\ \bottomrule
\end{tabular}}
\caption{Robustness of SC loss under different weight coefficients}
\label{different_ratios}
\end{table}

\subsubsection{Ablation study on various models}

We evaluate the effectiveness of our SC loss on light-weight networks including HRNet-W18-small~\cite{wang2020deep}, BiseNetV2~\cite{yu2021bisenet} and ConnectNet. As shown in table~\ref{ablation study}, SC Loss is effective across these networks, where the mIoU metric improves in HRNet-W18-small, BiseNetV2, and ConnectNet respectively. Through enhancing the connectivity of the connected components, the models obtain the better segmentation performance.

\begin{table}[ht]
\centering
{
\begin{tabular}{l|cc}
 \toprule 
Model  & mIoU & Pixel Acc \\ \toprule
HRNet-W18-small  & 93.0 &  97.2 \\
HRNet-W18-small + SCL  & \textbf{94.5} & 97.8\\ \midrule
BiseNetV2   & 85.8 & 94.2 \\
BiseNetV2 + SCL &\textbf{ 87.5} & 94.8 \\ \midrule
ConnectNet   & 94.1 & 97.6 \\
ConnectNet + SCL  & \textbf{94.6} & 97.6\\ 
\bottomrule
\end{tabular}}
\caption{Ablation study on light-weight networks}
\label{ablation study}
\end{table}

\subsubsection{Comparision with other SOTA losses}
In this section, we prove the superiority of SC loss over other state-of-the-art losses including lovasz loss~\cite{berman2018lovasz}, and RMI loss~\cite{zhao2019region}. These loss focus on different aspects of semantic segmentation like class imbalance and structural information. We conduct these experiments on HRNet-W18-small with learning rate being 0.5. For fair comparison, we set the coeffecient of the compound losses as 0.01 for all of the experiments.

As shown in Table \ref{losses}, the SC loss we propose outperforms other loss methods. The experiment with SC loss has the best score in mIoU and pixel accuracy. The evaluation result shows the SC loss has SOTA performance for portrait segmentation. 

\begin{table}[ht]
\centering
{
\begin{tabular}{l|ccc}
 \toprule 
Loss  & mIoU & Pixel Acc \\ \toprule
CE Loss (baseline) & 93.0 & 97.2\\
CE Loss + Lovasz Loss  & 93.0 & 97.2 \\
CE Loss + RMI Loss  & 94.3 & 97.7\\
CE Loss + SC Loss  & \textbf{94.5} & \textbf{97.8}\\ \bottomrule
\end{tabular}}
\caption{Comparision with SOTA losses}
\label{losses}
\end{table}

\begin{figure*}[t]
\centerline{\includegraphics[width=0.9\textwidth]{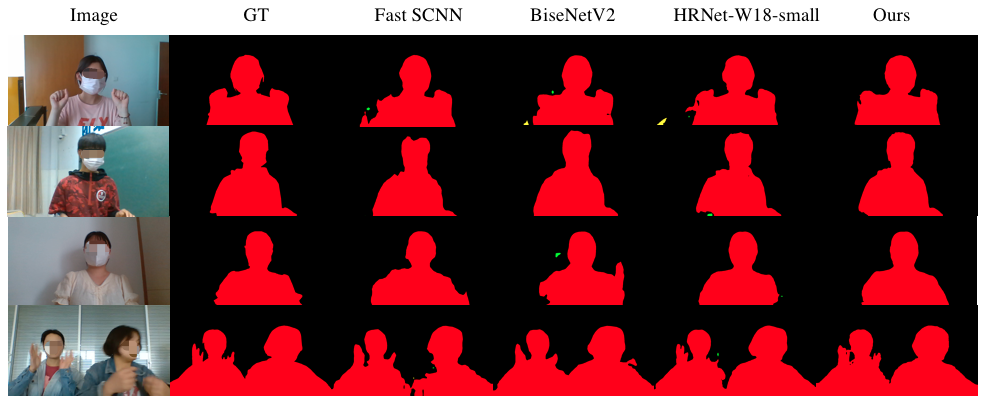}}
\caption{Semantic segmentation results of different light-weight networks}
\label{result}
\end{figure*}

\subsection{Effectiveness of ConnectNet}
In order to validate the performace of our proposed model, we compare its performance with other light-weight state-of-the-art models, including BiseNetV2~\cite{yu2021bisenet}, Fast SCNN~\cite{BMVC2019}, HRNet~\cite{wang2020deep}. 
As shown in Table \ref{benchmark}, our model is faster and more effective than other SOTA light-weight models. Compared with HRNet-W18-small, our model has greater performance and 41\% faster. Compared with Fast SCNN and BiseNetV2, our model is 1.5-3ms slower, but 1.2\% and 8.5\% higher in mIoU than BiseNetV2 and Fast SCNN, respectively.

The experimental results show that our model outperforms BiseNetV2 and Fast SCNN to a great extend, but only have less than 10\% of their parameters. This is crucial in mobile and web applications considering that the storage requirement is rather strict. 

\begin{table}[h!]
\centering
\resizebox{0.47\textwidth}{!}{
\begin{tabular}{l|ccccc}
 \toprule 
Model  & mIoU & Pixel Acc & Infer Time & Params\\ \toprule
BiseNetV2  & 85.8 & 94.2 & 10.0 & 2.32 \\
Fast SCNN  & 85.7 & 93.9 & \textbf{8.6} & 1.44 \\
HRNet-W18-small  & 93.0 & 97.2 & 19.76 & 3.95 \\
ConnectNet  & \textbf{94.2} & \textbf{97.6} & 11.5 & \textbf{0.13}\\ \bottomrule

\end{tabular}}
\caption{Benchmark on the state-of-the-art lightweight models. The unit of inference time is ms and the unit of Params is M.}
\label{benchmark}
\end{table}

\subsubsection{Qualitative Comparison}
In order to qualitatively show the performance of our network, We visualize the predictions of different networks on test images. As shown in figure \ref{result}, our model has better completeness than other models, and it is less prone to make disperse predictions.

\section{Conclusion}

To facilitate the progress in portrait segmentation in a video conferencing context, we introduce an open-source solution named PP-HumanSeg. In this work, we first construct a large-scale video portrait dataset that contains 291 videos from 23 conference scenes with 14K fine-labeled frames provided. To improve the completeness of segmentation results, we propose a Semantic Connectivity-aware Learning (SCL) framework incorporating a novel Semantic Connectivity (SC) loss. Such SC loss models the topology of portrait segmentation as a graph of connected components and measures the inconsistency between the graphs (i.e., connectivities) extracted from the ground truth labels and the prediction results as the loss. Furthermore, we propose an ultra-lightweight model, namely ConnectNet, with SCL for practical portrait segmentation. The proposed solution achieves the best trade-off between IoU and inference time in the dataset. Extensive evaluations on our dataset demonstrate the superiority of SCL and ConnectNet. The comparisons with other algorithms also show the advantage of proposed datasets from the perspectives of coverage and comprehensions. 

\section*{Acknowledgement}
This work was supported by the National Key Research and Development Project of China (2020AAA0103500).

{\small
\bibliographystyle{ieee_fullname}
\bibliography{egbib}
}

\end{document}